\documentclass[letterpaper, 10pt, conference]{ieeeconf}      

\IEEEoverridecommandlockouts                              

\usepackage{graphics} 
\usepackage{epsfig} 
\usepackage{subfigure}
\usepackage{amsmath} 
\usepackage{amssymb}  
\usepackage{epstopdf}
\usepackage{lipsum}
\usepackage{lipsum,amsmath,multicol}
\usepackage{float}
\usepackage{algorithm}
\usepackage{algpseudocode}
\usepackage{wrapfig}
\usepackage{sidecap}

\usepackage{amsthm}

\usepackage{comment}
\usepackage{array}
\usepackage{glossaries}
\makeglossaries
\newcolumntype{L}[1]{>{\raggedright\let\newline\\\arraybackslash\hspace{0pt}}m{#1}}
\newcolumntype{C}[1]{>{\centering\let\newline\\\arraybackslash\hspace{0pt}}m{#1}}
\newcolumntype{R}[1]{>{\raggedleft\let\newline\\\arraybackslash\hspace{0pt}}m{#1}}
\usepackage{sidecap}
\usepackage{url}
\usepackage[T1]{fontenc}
\usepackage{xcolor}
\usepackage{tabularx}
\usepackage{multicol, blindtext}
\usepackage{amsmath}

\allowdisplaybreaks

\makeatletter
\def\endthebibliography{%
  \def\@noitemerr{\@latex@warning{Empty `thebibliography' environment}}%
  \endlist
}
\makeatother

\begin{document}
%
\title{GPU Accelerated Convex Approximations for Fast Multi-Agent Trajectory Optimization}
\author{Fatemeh Rastgar, Houman Masnavi, Jatan Shrestha, Karl Kruusamäe, Alvo Aabloo, Arun Kumar Singh \thanks{All authors are with the Institute of Technology, University of Tartu. The work was supported in part by the European Social Fund through IT Academy program in Estonia, smart specialization project with BOLT and Estonian Centre of Excellence in IT (EXCITE) funded by the European Regional Development Fund.}
}
\maketitle



\begin{abstract}
In this paper, we present a computationally efficient trajectory optimizer that can exploit GPUs to jointly compute trajectories of tens of agents in under a second. At the heart of our optimizer is a novel reformulation of the non-convex collision avoidance constraints that reduces the core computation in each iteration to that of solving a large scale, convex, unconstrained Quadratic Program (QP). We also show that the matrix factorization/inverse computation associated with the QP needs to be done only once and can be done offline for a given number of agents. This further simplifies the solution process, effectively reducing it to a problem of evaluating a few matrix-vector products. Moreover, for a large number of agents, this computation can be trivially accelerated on GPUs using existing off-the-shelf libraries. We validate our optimizer's performance on challenging benchmarks and show substantial improvement over state of the art in computation time and trajectory quality.

\end{abstract}

\section{Introduction}
Coordinating multiple agents between given start and goal positions without collision is crucial to any multi-agent application. For highly agile agents like quadrotors and autonomous cars, a popular approach has been to formulate this collision-free coordination as a trajectory optimization problem. There are two core computational challenges in this context. First, the number of variables in the optimization problem increases linearly with the number of agents $n$. Second and more importantly, the number of pair-wise non-convex collision avoidance constraints grow by a factor $\binom{n}{2}$.  Existing works have predominantly explored two classes of simplifications to keep the optimization problem tractable. The sequential approaches, e.g., \cite{iscp}, \cite{rbp_quad}, follow an iterative process wherein motion plans for only one agent is computed at a time. Collision avoidance is ensured by treating agents whose motions were computed in earlier iterations as dynamic non-responsive obstacles for the currently planned agent. On the other hand, the distributed model predictive control (MPC) approaches such as \cite{luis2019_dmpc}, \cite{dmpc_alonso_mora} decouple the planning process by allowing each agent to view all the others at any given instant as dynamic obstacles with a known trajectory. The sequential approaches do not leverage the cooperation between the agents while it is incorporated only implicitly (through trajectory prediction) in the distributed MPC based approaches. As a result, both class of simplifications have access to a smaller feasible space and are thus conservative. 

In this paper, we explicitly account for inter-agent cooperation by adopting the classical set-up of \cite{rafealla_ccp_quad5}, wherein a large scale optimization is formulated to compute the trajectories of all the agents jointly. The primary goal of this paper is to improve the computational tractability of such large scale optimization problems. Our optimizer falls into the class of existing algorithms such as \cite{alonso_nips} that reformulates the underlying numerical computation of the optimizer to parallelize them over CPUs/GPUs.  

\subsection{Main Idea}
\noindent Consider the following unconstrained, quadratic program (QP) for a constant matrix $\textbf{Q}$ and a vector $\textbf{q}$. As shown, the solution process reduces to solving a set of linear equations.

\begin{align}
    \min_{\boldsymbol{\xi}} \frac{1}{2}\boldsymbol{\xi}^T\textbf{Q} \boldsymbol{\xi}+\textbf{q}^T\boldsymbol{\xi}, \label{cost_ex_qp} \Rightarrow  \textbf{Q} \boldsymbol{\xi} = -\textbf{q}
    \end{align}




\noindent Now, imagine that QP (\ref{cost_ex_qp}) needs to be solved for several instantiations of \textbf{q} for a given  $\textbf{Q}$. Such scenarios are common in linear MPC, wherein $\textbf{Q}$ encodes the system dynamics and cost functions and is thus constant, while $\textbf{q}$ that encodes the initial condition changes at each iteration. As shown in \cite{boyd_os_mpc}, an efficient way of handling such scenarios is to pre-compute the inverse (or just the factorization) of   $\textbf{Q}$, in which case, MPC computation (or solving (\ref{cost_ex_qp})) in each iteration reduces to evaluating just matrix-vector products. This reduction becomes particularly important in cases where the QP (\ref{cost_ex_qp}) is formulated over tens of hundreds of variables. 

\subsection{Contributions}
\noindent For the first time, we show how the idea of off-line caching of matrix inverses can be used to accelerate non-linear and non-convex, multi-agent trajectory optimization problems. This is achieved by deriving the multi-agent version of the collision avoidance constraints presented in our prior work \cite{aks_iros20}. Subsequently, we use Alternating Minimization \cite{jain_mlbook} to reduce the trajectory optimization into smaller sub-problems wherein the most computationally intensive block has a structure similar to (\ref{cost_ex_qp}). It is important to note that although QP based approaches are exceedingly common in multi-agent trajectory optimization, their mathematical structure precludes leveraging pre-computed matrix inverses/factorization. (see Remark \ref{remark_3} in Section \ref{analysis}). Our optimizer provides the following benefits over the current state of the art.



\noindent  \textbf{Ease of Implementation and GPU Accelerations:} The entire numerical computation of our optimizer reduces to computing either element-wise operations over vectors or matrix-vector products. For a large number of agents, these can be trivially accelerated on GPUs using libraries like CUPY \cite{cupy} and JAX \cite{jax}. We also provide an open source implementation in \url{https://github.com/arunkumar-singh/GPU-Multi-Agent-Traj-Opt}. Our GPU accelerated optimizer can compute trajectories for 32 agents in 0.7 s on a RTX-2080 enabled desktop computer.  

\noindent \textbf{State of the Art Performance:} Our optimizer outperforms the computation time of joint trajectory optimization of \cite{rafealla_ccp_quad5} by several orders of magnitude while achieving trajectories of similar quality. It also outperforms the current state of the art, sequential approach of \cite{rbp_quad}, by obtaining shorter trajectories in all the considered benchmarks. Even more importantly, our optimizer also outperforms \cite{rbp_quad} in terms of the computation time on several benchmarks. The speed-up is particularly impressive given that our optimizer performs a much more rigorous joint search over the agents' trajectory space.

\noindent \textbf{Suitability on Edge-Devices:} Our optimizer can compute trajectories of 16 agents in around $2s$ on Nvidia Jetson-TX2. This is orders of magnitude faster than the computation time of \cite{rafealla_ccp_quad5} on an Intel i7 desktop computer with 32GB RAM. Thus, our work presents an important step towards achieving complex onboard decision-making abilities for light-weight quadrotors.

\section{Background and Preliminaries}
This section introduces some necessary mathematical preliminaries and uses them to draw a contrast between existing works and our optimizer. We begin by summarizing next the basic symbols and notations used throughout the paper.

\subsection{Symbols and Notations}
\noindent We will use lower case normal font letters to represent scalars, while bold font variants represent vectors. Matrices are represented through upper case bold fonts. The time dependency of the variable is shown by $t$. The superscript $T$ will denote the transpose of vectors and matrices. The left superscript $k$ will be used to indicate the iteration index in a trajectory optimizer. We use subscript $i, j$ as agent index. We will use $n, m$ to represent the number of agents and planning horizon throughout the paper.

\subsection{Quadratic Inequalities and Conservative Convex Bounds }
\noindent For spheroid agents with dimension $l_{xy}, l_z$, the inter-agent collision avoidance constraints take the following non-convex quadratic inequality form.

\small
\begin{align}
    f_{c}(x_i(t), y_i(t), z_i(t), x_j(t), y_j(t), z_j(t)) = \nonumber \\ -\frac{(x_i(t)-x_j(t))}{l_{xy}^2}-\frac{(y_i(t)-y_j(t))}{l_{xy}^2}-\frac{(z_i(t)-z_j(t))}{l_z^2}+1 \leq 0, \label{quad_multiagent}
\end{align}
\normalsize

\noindent where, $(x_i(t), y_i(t), z_i(t))$ represent the position of the $i^{th}$ at some time $t$. Interestingly, a simple linearization of (\ref{quad_multiagent}) around any arbitrary guess trajectory leads to a convex but conservative approximation of the feasible space \cite{boyd_ccp}, \cite{shen_ccp_quad_1}. This simplification has been exploited in many recent works on multi-agent trajectory optimization such as \cite{rafealla_ccp_quad5}, \cite{iscp}. However, the conservativeness often leads to an infeasible optimization problem even when a solution exists. To side-step this bottleneck, \cite{iscp} relaxes the trajectory optimization by only incrementally enforcing the collision avoidance constraints.

Our optimizer induces convexity in a uniquely different way based on our prior work \cite{aks_iros18}, \cite{aks_icra20}, \cite{aks_ecc20}. Instead of relying on linearization, it breaks down the problem into smaller sub-problems where all but one are convex optimization problems. Moreover, the non-convex sub-problem has a geometrical structure that allows for obtaining an approximate analytical solution.

\subsection{Joint Trajectory Optimization of \cite{rafealla_ccp_quad5} }
\noindent Reference \cite{rafealla_ccp_quad5} employs a sequential convex programming (SCP) based optimizer on the multi-agent trajectory optimization problem. Let $({^k}x_i(t), {^k}y_i(t), {^k}z_i(t) )$ be the solution guess at iteration $k$. Then, the SCP of \cite{rafealla_ccp_quad5} solves the following QP at iteration $k+1$.

\small
\begin{subequations}
\begin{align}
    \min_{x_i(t), y_i(t), z_i(t)} \sum_i \sum_t \ddot{x}^{2}_i(t)+\ddot{y}^{2}_i(t)+\ddot{z}^{2}_i(t) \label{cost_multiagent}  \\
    (x_i(t), y_i(t), z_i(t)) \in \mathcal{C}_{boundary} \label{eq_multiagent}\\
    {^k}\textbf{A}_{ij}\begin{bmatrix}
    x_i(t)\\
    y_i(t)\\
    x_j(t)\\
    y_j(t)
    \end{bmatrix}\leq  {^k}\textbf{b}_{ij}, \forall t, i, j, i\neq j \label{coll_multiagent}
\end{align}
\end{subequations}
\normalsize

\noindent where, ${^k}\textbf{A}_{ij}, {^k}\textbf{b}_{ij}$ are obtained by linearization of (\ref{quad_multiagent}) around $({^k}x_i(t), {^k}y_i(t), {^k}z_i(t) )$. The constraints (\ref{eq_multiagent}) force the initial and final boundary conditions of the agents to lie in the set $\mathcal{C}_{boundary}$. The main complexity of QP (\ref{cost_multiagent})-(\ref{coll_multiagent}) stems from the fact the number of affine inequality constraints increases by a factor of $n\choose 2$. As mentioned earlier, works like \cite{iscp} by-pass this intractability by adopting a sequential approach.

Similarly to \cite{rafealla_ccp_quad5}, our optimizer also reduces to solving QPs over the joint trajectory space of all the agents. However, it does not have any inequality constraints. Furthermore, the most expensive part of the solution process can be pre-computed for a given number of agents.

\subsection{GPU Acceleration Through Gradient Descent}
\noindent An effective way of accelerating optimization problems on GPU is to reformulate them in an unconstrained form and then apply the method of Gradient Descent. For example, \cite{quad_gd_gpu} achieves this for the multi-agent trajectory optimization by augmenting the constraints (\ref{eq_multiagent})-(\ref{coll_multiagent}) as penalties in the cost function (\ref{cost_multiagent}). 




The core computations in Gradient Descent reduces to computing matrix-vector products which can be readily accelerated on GPUs. Reference \cite{quad_gd_gpu} essentially exploits this feature and also brings in additional innovation in terms of parallelizing collision checks to further improve the computation time.

Our optimizer provides substantial improvements over \cite{quad_gd_gpu}. As mentioned by the authors themselves, \cite{quad_gd_gpu} requires extensive hyper-parameter tuning that is likely to be redone if the problem parameters such as robot dimension change. In contrast, the proposed optimizer relies on accelerating a QP on GPU by offline caching of matrix inverses and worked with trivial default parameters on dozens of examples.

\section{Main Results}

\noindent In this section, we present our main theoretical result: a GPU accelerated optimizer based on convex optimization. We begin by reiterating the main assumptions.

\begin{itemize}
    \item We consider agents with decoupled affine motion models along the $(x,y, z)$  axis. We also assume differential flatness which allows us to extract control inputs from position derivatives. This is typical of holonomic agents like quadrotors \cite{rafealla_ccp_quad5}, \cite{iscp}. Even autonomous cars  can be modeled in this form under some conditions \cite{miqp_planner}.  
    
    \item The agents are modeled as spheroids (or disks in 2D).
    
    \item We do not explicitly consider the bounds on velocities and accelerations and rely on choosing an appropriate traversal time and regularization on accelerations to ensure the same. However, it is possible to reformulate bounds as quadratic penalties and incorporate within the optimizer without disturbing its computational structure (see eqn (18) in \cite{admm_qp}). Alternately, like \cite{rbp_quad}, we can also scale the traversal time during post-processing to satisfy the bounds.
    
\end{itemize}


\subsection{Reformulation and Alternating Minimization}
\noindent We reformulate (\ref{cost_multiagent})-(\ref{coll_multiagent}) in the following form. 


\small
\begin{subequations}
\begin{align}
    \min_{x_i, y_i, z_i, \alpha_{ij}, \beta_{ij}, d_{ij}}  \sum_i \sum_t \ddot{x}^{2}_i(t)+\ddot{y}^{2}_i(t)+\ddot{z}^{2}_i(t) \label{cost_proposed}  \\
        (x_i(t), y_i(t), z_i(t)) \in \mathcal{C}_{boundary} \label{eq_proposed}\\
    \textbf{f}_{c} = \textbf{0}, \forall i, j, t \label{coll_proposed}\\
    \beta_{ij}(t) \in [0, \pi],  \alpha_{ij}(t) \in [-\pi, \pi], d_{ij}(t)\geq 1, \forall i,j, t \label{alpha_beta_constr}
\end{align}
\end{subequations}

\begin{align}
    \textbf{f}_{c} = \left \{ \begin{array}{lcr}
x_i(t) -x_j(t)-l_{xy}d_{ij}(t)\sin\beta_{ij}(t)\cos\alpha_{ij}(t) \\
y_i(t) -y_j(t)-l_{xy}d_{ij}(t)\sin\beta_{ij}(t)\sin\alpha_{ij}(t)\\ 
z_i(t) -z_j(t)-l_z d_{ij}\cos\beta_{ij}(t) \\
\end{array} \right \}
\label{sphere_proposed}
\end{align}
\normalsize

\noindent The primary changes involve introducing additional time-dependent variables $\alpha_{ij}(t), \beta_{ij}(t), d_{ij}(t)$, and using them to rephrase quadratic collision avoidance constraints (\ref{quad_multiagent}) into a set of non-linear equalities (\ref{sphere_proposed}). On the surface, our formulation (\ref{cost_proposed})-(\ref{alpha_beta_constr}) looks more complicated than the more conventional multi-agent trajectory optimization (\ref{cost_multiagent})-(\ref{coll_multiagent}) as the former involves highly non-linear trigonometric functions. But in fact, (\ref{cost_proposed})-(\ref{alpha_beta_constr}) has some hidden geometrical and computational structures that we can expose using techniques from Alternating Minimization (AM). To this end, we first create an augmented cost function $\mathcal{L}$ by incorporating $\textbf{f}_{c}$ as $l_2$ penalties 

\small
\begin{subequations}
\begin{align}
    \mathcal{L} = \sum_{i, t}  \ddot{x}^{2}_i(t)+\ddot{y}^{2}_i(t)+\ddot{z}^{2}_i(t)+ \nonumber \\
    \sum_{i, j, t} \frac{\rho}{2}(x_i(t) -x_j(t)-l_{xy}d_{ij}(t)\sin\beta_{ij}(t)\cos\alpha_{ij}(t)+\frac{\lambda_{xij}(t)}{\rho})^2 \nonumber \\
    +\frac{\rho}{2}(y_i(t) -y_j(t)-l_{xy}d_{ij}(t)\sin\beta_{ij}(t)\sin\alpha_{ij}(t)+\frac{\lambda_{yij}(t)}{\rho})^2 \nonumber \\
    +\frac{\rho}{2}(z_i(t) -z_j(t)-l_{z}d_{ij}(t)\cos\beta_{ij}(t)+\frac{\lambda_{zij}(t)}{\rho})^2
    \label{aug_lagrange}
\end{align}
\end{subequations}
\normalsize

\noindent In (\ref{aug_lagrange}), $\rho$ is a scalar constant and $\lambda_{xij}(t), \lambda_{yij}(t), \lambda_{zij}(t)$ are time-dependent Lagrange multipliers that can be used to drive the residual of $\textbf{f}_{c}$ to zero. Algorithm \ref{algo_1} summarizes the minimization of (\ref{aug_lagrange}) subject to (\ref{eq_proposed}) and (\ref{alpha_beta_constr}) based on the AM technique. As before, the left superscript $k$ represents the respective variable at iteration $k$. As shown, we start with an initialization for ${^k}\alpha_{ij}(t), {^k}\beta_{ij}(t), {^k}d_{ij}(t), {^k}\lambda_{xij}(t), {^k}\lambda_{yij}(t), {^k}\lambda_{zij}(t) $ at $k = 0$ and optimize the variables in a sequence. The optimizations (\ref{step_x_3d})-(\ref{step_z_3d}) and (\ref{step_d}) are convex constrained QPs. In contrast, (\ref{step_alpha}) is non-convex. But interestingly, simple geometrical intuitions can be used to derive an analytical solution for it. We delve deeper into each of these optimizations next.

\subsection{Analysis of Algorithm \ref{algo_1} } \label{analysis}
\noindent \subsubsection{Steps (\ref{step_x_3d})-(\ref{step_z_3d}) } The most important feature of these optimizations is that each of them takes the form of a QP wherein the matrices do not change over iteration. To validate this assertion and to show how it is useful, we parametrize $x_i(t)$ and its derivatives in the following form.

\begin{equation}
\begin{bmatrix}
x_i(t_1)\\
x_i(t_2)\\
\dots\\
x_i(t_n)
\end{bmatrix} = \textbf{P}\textbf{c}_{x_i}, \begin{bmatrix}
\dot{x}_i(t_1)\\
\dot{x}_i(t_2)\\
\dots\\
\dot{x}_i(t_n)
\end{bmatrix} = \dot{\textbf{P}}\textbf{c}_{x_i},  \begin{bmatrix}
\ddot{x}_i(t_1)\\
\ddot{x}_i(t_2)\\
\dots\\
\ddot{x}_i(t_n)
\end{bmatrix} = \ddot{\textbf{P}}\textbf{c}_{x_i},
\label{param}
\end{equation}

\noindent where, $\textbf{P}$ is a matrix formed with time-dependent basis functions (e.g polynomials) and $\textbf{c}_{x_i}$ are the coefficients associated with the basis functions. Let $\textbf{c}_x$ be the joint coefficient formed by stacking $\textbf{c}_{x_i}$ for all the agents. Now, with the help of (\ref{param}), we can derive the following matrix representation for optimization (\ref{step_x_3d})

\begin{subequations}
\begin{align}
\min \frac{1}{2}\textbf{c}_x^T(\textbf{Q}_x+\rho\textbf{A}_{f_c}^T\textbf{A}_{f_c}) \textbf{c}_x+(-\textbf{A}_{f_c}^T{^k}\textbf{b}_{f_c}^x)^T\textbf{c}_x    \label{cost_step_x}\\
\textbf{A}_{eq}\textbf{c}_x = \textbf{b}_{eq}^x \label{eq_step_x}
\end{align}
\end{subequations}

\noindent Note, how only the vector ${^k}\textbf{b}_{f_c}^x$ is shown to be dependent on the iteration index $k$. The various matrices and vectors involved in (\ref{cost_step_x})-(\ref{eq_step_x}) are derived in the following manner.

\begin{align}
    \sum_{i,t}\ddot{x}_i(t)^2 \Rightarrow \frac{1}{2}\textbf{c}_x^T\textbf{Q}_x\textbf{c}_x,  \textbf{Q}_x = \begin{bmatrix}
    \ddot{\textbf{P}}^T\ddot{\textbf{P}} & & \\
    & \ddots  & \\
    & & \ddot{\textbf{P}}^T\ddot{\textbf{P}}
    \end{bmatrix}
\end{align}

\begin{align}
    x_i(t) \in \mathcal{C}_{boundary}, \forall i \Rightarrow \textbf{A}_{eq} \textbf{c}_x = \textbf{b}_{eq}^x
\end{align}

\begin{align}
    \textbf{A}_{eq} = \begin{bmatrix}
    \textbf{A} & & \\
    & \ddots & \\
    & & \textbf{A}
    \end{bmatrix}, \textbf{A} = \begin{bmatrix}
    \textbf{P}_1\\
    \dot{\textbf{P}}_1\\
    \ddot{\textbf{P}}_1\\
    \textbf{P}_m\\
    \dot{\textbf{P}}_m\\
    \ddot{\textbf{P}}_m\\
    \end{bmatrix}
\end{align}

\noindent The matrix $\textbf{A}$ is formed by stacking the first and last row of $\textbf{P}$ and its derivatives, and $\textbf{b}_{eq}^x$ is formed by stacking initial and final position, velocities and accelerations. 

\noindent Similarly, we obtain the following matrix representation. 

\small
\begin{align}
\sum_{i, j, t} \frac{\rho}{2}(x_i(t) -x_j(t)-l_{xy}{^k}d_{ij}(t)\sin{^k}\beta_{ij}(t)\cos{^k}\alpha_{ij}(t)+\frac{{^k}\lambda_{xij}(t)}{\rho})^2  \nonumber \\
\Rightarrow \frac{\rho}{2}\Vert \textbf{A}_{f_c} \textbf{c}_x-{^k}\textbf{b}_{f_c}^x\Vert_2^2, \textbf{A}_{f_c} =  \begin{bmatrix}
    \textbf{A}_{1} & & \\
    & \ddots & \\
    & & \textbf{A}_{r}
  \end{bmatrix}
\end{align}
\normalsize


\begin{align}
    \textbf{A}_r = \begin{bmatrix}
     \begin{pmatrix}
     \textbf{P}\\
     \textbf{P}\\
      \vdots\\
      \textbf{P}
     \end{pmatrix}_{\times n-r}&\begin{bmatrix}
    -\textbf{P} & & \\
    & \ddots & \\
    & & -\textbf{P}
  \end{bmatrix}_{\times n-r}
    \end{bmatrix}
    \label{A_f}
\end{align}

\begin{align}
    {^k}\textbf{b}_{f_c}^x = l_{xy}{^k}\textbf{d} \sin{^k}\boldsymbol{\beta}\cos{^k}\boldsymbol{\alpha}-\frac{\boldsymbol{{^k}\lambda_{xij}}}{\rho}
    \label{b_f}
\end{align}

\noindent In (\ref{A_f}), the operator $(.)_{\times n-r}$ vertically stacks the matrix $\textbf{P}$ $n-r$ times. Similarly, the block-diagonal matrix on the r.h.s is formed by $(n-r)$ number of matrix \textbf{P}. In (\ref{b_f}), ${^k}\textbf{d}$, ${^k}\boldsymbol{\beta}$, ${^k}\boldsymbol{\alpha}$ are formed by stacking the respective variables at all times and for all the agents. Similar construction is also followed for ${^k}\boldsymbol{\lambda}_{xij}$.

\begin{algorithm*}[!h]
 \caption{Alternating Minimization based Multi-Agent Trajectory Optimization }\label{algo_1}
    \begin{algorithmic}[1]   
\State Initialize   ${^k}d_{ij}(t), {^k}\alpha_{ij}(t), {^k}\beta_{ij} $ at $k = 0$
\While {$k\leq maxiter$ \text{ or till norm of the residuals are below some threshold}}

\begin{subequations}
\small
\begin{align}
{^{k+1}}x_i(t) = \arg\min_{x_i(t)\in \mathcal{C}_{boundary}} \sum_{i, t}\ddot{x}_i(t)^2+\sum_{i, j, t}\frac{\rho}{2}(x_i(t) -x_j(t)-l_{xy}{^k}d_{ij}(t)\sin{^k}\beta_{ij}(t)\cos{^k}\alpha_{ij}(t)+\frac{{^k}\lambda_{xij}(t)}{\rho})^2 \label{step_x_3d}\\ 
{^{k+1}}y_i(t) = \arg\min_{y_i(t)\in \mathcal{C}_{boundary}} \sum_{i, t}\ddot{y}_i(t)^2+\sum_{i, j, t}\frac{\rho}{2}(y_i(t) -y_j(t)-l_{xy}{^k}d_{ij}(t)\sin{^k}\beta_{ij}(t)\sin{^k}\alpha_{ij}(t)+\frac{{^k}\lambda_{yij}(t)}{\rho})^2
 \label{step_y_3d} \\
{^{k+1}}z_i(t) = \arg\min_{z_i(t)\in \mathcal{C}_{boundary}} \sum_{i, t}\ddot{z}_i(t)^2+\sum_{i, j, t}\frac{\rho}{2}(z_i(t) -z_j(t)-l_{z}{^k}d_{ij}(t)\cos{^k}\beta_{ij}(t)+\frac{{^k}\lambda_{zij}(t)}{\rho})^2
 \label{step_z_3d}\\
 {^{k+1}}\alpha_{ij}(t), {^{k+1}}\beta_{ij}(t) = \arg\min_{\alpha_{ij}, \beta_{ij}} \sum_{i, j, t} \frac{\rho}{2}({^{k+1}}x_i(t) -{^{k+1}}x_j(t)-l_{xy}{^k}d_{ij}(t)\sin\beta_{ij}(t)\sin\alpha_{ij}(t)+\frac{{^k}\lambda_{xij}(t)}{\rho})^2\nonumber \\
 +\frac{\rho}{2}({^{k+1}}y_i(t) -{^{k+1}}y_j(t)+l_{xy}{^k}d_{ij}(t)\sin\beta_{ij}(t)\cos\alpha_{ij}(t)+\frac{{^k}\lambda_{yij}(t)}{\rho})^2 \nonumber \\
 +\frac{\rho}{2}({^{k+1}}z_i(t) -{^{k+1}}z_j(t)+l_{z}{^k}d_{ij}(t)\cos\beta_{ij}(t)+\frac{{^k}\lambda_{zij}(t)}{\rho})^2 \label{step_alpha}\\
 \approx   {^{k+1}}\alpha_{ij}(t) = \arctan2({^{k+1}}y_i(t)-{^{k+1}}y_j(t), {^{k+1}}x_i(t)-{^{k+1}}x_j(t)) \nonumber \\
 {^{k+1}}\beta_{ij} = \arctan2(\frac{{^{k+1}}x_i(t)-{^{k+1}}x_j(t)}{l_{xy} \cos{{^{k+1}}\alpha_{ij}(t)}}, \frac{{^{k+1}}z_i(t)-{^{k+1}}z_j(t)}{l_z} ) \label{sphere_project} \\
 {^{k+1}}d_{ij}(t) = \arg\min_{d_{ij}} \sum_{i, j, t}\frac{\rho}{2}({^{k+1}}x_i(t) -{^{k+1}}x_j(t)-l_{xy}d_{ij}(t)\sin{^{k+1}}\beta_{ij}(t)\cos{^{k+1}}\alpha_{ij}(t)+\frac{{^{k}}\lambda_{xij}(t)}{\rho})^2 \nonumber \\
 +\frac{\rho}{2}({^{k+1}}y_i(t) -{^{k+1}}y_j(t)-l_{xy}d_{ij}(t)\sin{^{k+1}}\beta_{ij}(t)\sin{^{k+1}}\alpha_{ij}(t)+\frac{{^{k}}\lambda_{yij}(t)}{\rho})^2\nonumber \\
 +\frac{\rho}{2}({^{k+1}}z_i(t) -{^{k+1}}z_j(t)-l_{z}d_{ij}(t)\cos{^{k+1}}\beta_{ij}(t)+\frac{{^{k}}\lambda_{zij}(t)}{\rho})^2 \label{step_d} \\
 {^{k+1}}\lambda_{xij}(t) =  {^{k}}\lambda_{xij}(t)+\rho({^{k+1}}x_i(t) -{^{k+1}}x_j(t)-l_{xy}{^{k+1}}d_{ij}(t)\sin{^{k+1}}\beta_{ij}(t)\cos{^{k+1}}\alpha_{ij}(t)) \nonumber \\
 {^{k+1}}\lambda_{yij}(t) = {^{k}}\lambda_{yij}(t)+\rho({^{k+1}}y_i(t) -{^{k+1}}y_j(t)-l_{xy}{^{k+1}}d_{ij}(t)\sin{^{k+1}}\beta_{ij}(t)\sin{^{k+1}}\alpha_{ij}(t)) \nonumber \\
 {^{k+1}}\lambda_{zij}(t) =  {^{k}}\lambda_{zij}(t)+\rho({^{k+1}}z_i(t) -{^{k+1}}z_j(t)-l_{z}{^{k+1}}d_{ij}(t)\cos{^{k+1}}\beta_{ij}(t)) \label{lag_update}
\end{align}
\end{subequations}
\normalsize
\EndWhile
\end{algorithmic}  
\end{algorithm*}

\noindent \textbf{Reduction to Linear Equations:} The equality constrained QP (\ref{cost_step_x})-(\ref{eq_step_x}) can be reduced to a problem of solving a set of linear equations \cite{boyd_os_mpc}.

\begin{align}
    \overbrace{\begin{bmatrix}
    (\textbf{Q}_x+\rho\textbf{A}_{f_c}^T\textbf{A}_{f_c}) & \textbf{A}_{eq}^T\\
    \textbf{A}_{eq} & \textbf{0}
    \end{bmatrix} }^{\widetilde{\textbf{Q}}_x} \textbf{c}_x = \overbrace{\begin{bmatrix}
    \textbf{A}_{f_c}^T{^k}\textbf{b}_{f_c}^x\\
    \textbf{b}_{eq}^x
    \end{bmatrix}}^{\widetilde{\textbf{q}}_x}
\label{kkt_step_x}
\end{align}

\newtheorem{remark}{Remark}
\begin{remark}\label{remark_1}
For a given $\rho$, the matrix on the l.h.s of (\ref{kkt_step_x}) is independent of the iteration index $k$. Thus, its inverse can be pre-computed and used without any computational cost in each iteration of Algorithm \ref{algo_1}.
\end{remark}

\begin{remark}\label{remark_2}
For a given $\rho$ and the number of agents $n$, the matrix on the l.h.s of (\ref{kkt_step_x}) only depends on our choice of trajectory parametrization $\textbf{P}$. Thus, the same pre-computed inverse can be used to solve trajectory optimization for any variations of start and goal positions as long as the number of agents and trajectory parametrization remains the same.
\end{remark}

\begin{remark}\label{remark_3}
It is possible to reformulate optimization (\ref{cost_multiagent})-(\ref{coll_multiagent}) used in works like \cite{rafealla_ccp_quad5}, \cite{iscp}, \cite{luis2019_dmpc} in the form of equality constrained QP and subsequently reduce it a problem of linear equation solving. However, it will not be possible to pre-compute the matrix inverses in this approach because the matrix ${^k}\textbf{A}_{ij}$ in (\ref{coll_multiagent}) will change at each iteration of the sequential convex programming optimizer. 
\end{remark}

\noindent{\textbf{Per-Iteration Complexity}:} Let, $n_v$ be the number of decision variables of each agent (columns of \textbf{P}). Let $n_b$ be the number of boundary conditions (row of $\textbf{A}$) for each agent along each motion axis. Then the per-iteration complexity of step (\ref{step_x_3d}) or (\ref{kkt_step_x}) is dominated by two large matrix-vector products. The first product involves multiplying $\textbf{A}_{f_c}^T$ with dimensions $(nn_v \times m {n\choose 2})$ with vector ${^k}\textbf{b}_{f_c}^x$ of $m {n\choose 2} \times 1$, where we recall $n, m$ to be the number of agents and length of the planning horizon respectively. The complexity of second matrix-vector product, $\widetilde{\textbf{Q}}_x^{-1}\widetilde{\textbf{q}}_x$ is $\mathcal{ O }(n^2(n_b+n_v)^2)$ and follows similar reasoning.  However, it should be noted that these are worst-case complexities without the GPU parallelization. For example, theoretically, $\textbf{A}_{f_c}^T {^k}\textbf{b}_{f_c}^x$ can be split into $nn_v$ parallel computations. However, in practice, the speed-up through parallelization depends on the  number of available GPU cores and other hardware limitations such as speed of CPU-GPU transfer.

\noindent The above analysis drawn for (\ref{step_x_3d}) can be trivially extended to steps (\ref{step_y_3d})-(\ref{step_z_3d}) as well.

\noindent \subsubsection{Step (\ref{step_alpha})} 
\begin{figure}[!h]
  \centering  
    \includegraphics[width= 7.5cm, height=5.0cm] {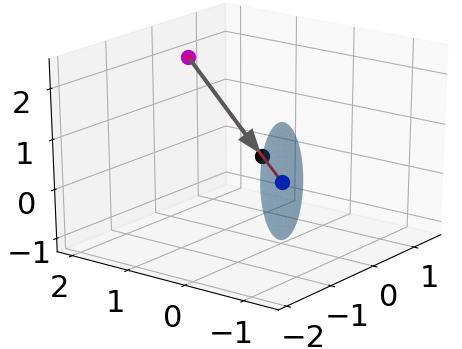}
\caption{Graphical description of step (\ref{sphere_project}). The magenta colored point corresponds to $({^{k+1}}x_i(t)-{^{k+1}}x_j(t))$, $({^{k+1}}y_i(t)-{^{k+1}}y_j(t))$, $({^{k+1}}z_i(t)-{^{k+1}}z_j(t))$. Step (\ref{sphere_project}) projects this point onto a spheroid centered at origin. The projected point is shown in black and the center of the spheroid is shown in blue.}
\label{projection_def}
\end{figure}

\noindent Although optimization (\ref{step_alpha}) is non-convex, an approximate solution can be derived using simple geometrical intuition. Recall (\ref{sphere_proposed}) to note that the set of feasible $(x_i(t)-x_j(t))$, $(y_i(t)-y_j(t))$ and $(z_i(t)-z_j(t))$ constitute a spheroid centered at origin with dimensions $l_{xy}d_{ij}$ and $l_{z}d_{ij}$. Thus, we compute ${^{k+1}}\alpha_{ij}(t)$ and ${^{k+1}}\beta_ij(t)$ by projecting  ${^{k+1}x_i(t)}$, ${^{k+1}y_i(t)}$, ${^{k+1}z_i(t)}$ obtained from steps (\ref{step_x_3d})-(\ref{step_z_3d}) onto the requisite spheroid through equations (\ref{sphere_project}). The projection process is also illustrated in Fig.\ref{projection_def}. The projection satisfies the constraints on $\alpha_{ij}(t)$ by construction. For $\beta_{ij}(t)$, we simply clip the values to $[0, \pi]$. 

\noindent \subsubsection{Step \ref{step_d}} For a given pair of agents $(i, j)$, $d_{ij}(t)$ at different time instants are decoupled from each other. Similarly, they are also decoupled across agent pairs. Thus, (\ref{step_d}) splits into $m*{n\choose 2}$ decoupled single-variable convex QPs, each of which can be solved symbolically. That is the solutions are available as analytical formulae that we can evaluate using just element wise operations over vectors. The constraints on $d_{ij}(t)$ are ensured by simply clipping the values to $[0 \hspace{0.1cm} 1]$ at each iteration.


\noindent \subsubsection{Step \ref{lag_update} } These steps update the Lagrange multipliers based on the residuals achieved at the current iteration \cite{boyd_admm}.


\begin{figure*}[!h]
  \centering  
  \subfigure[]{
    \includegraphics[width= 5.7cm, height=5.0cm] {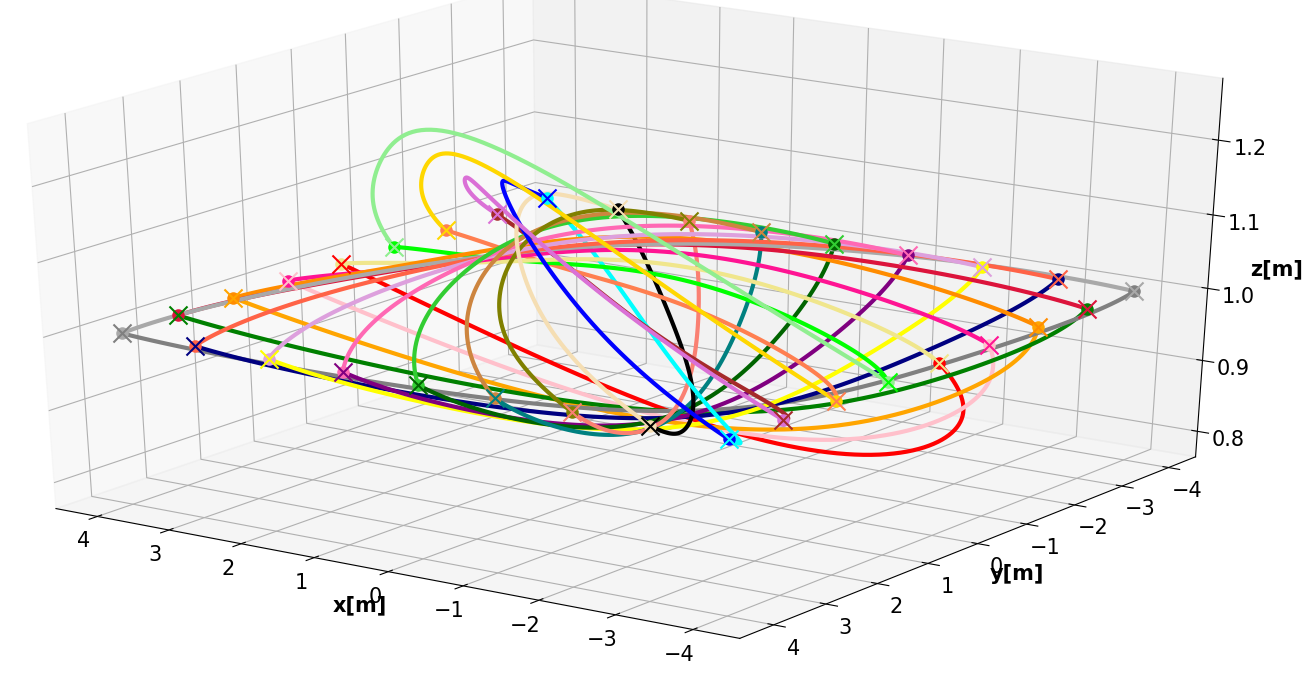}
    \label{square_benchmark}
   }\hspace{-0.4cm}
\subfigure[]{
    \includegraphics[width= 5.7 cm, height=5.0cm] {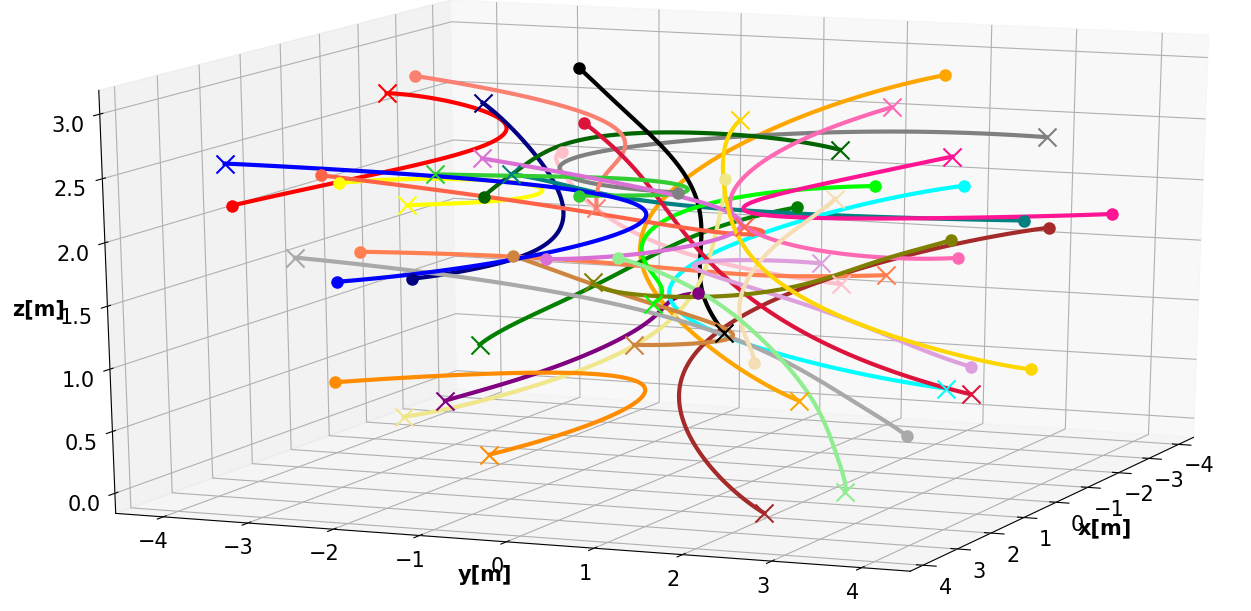}
    \label{random_benchmark}
   }\hspace{-0.4cm}
  \subfigure[]{
    \includegraphics[width= 5.7cm, height=5.0cm] {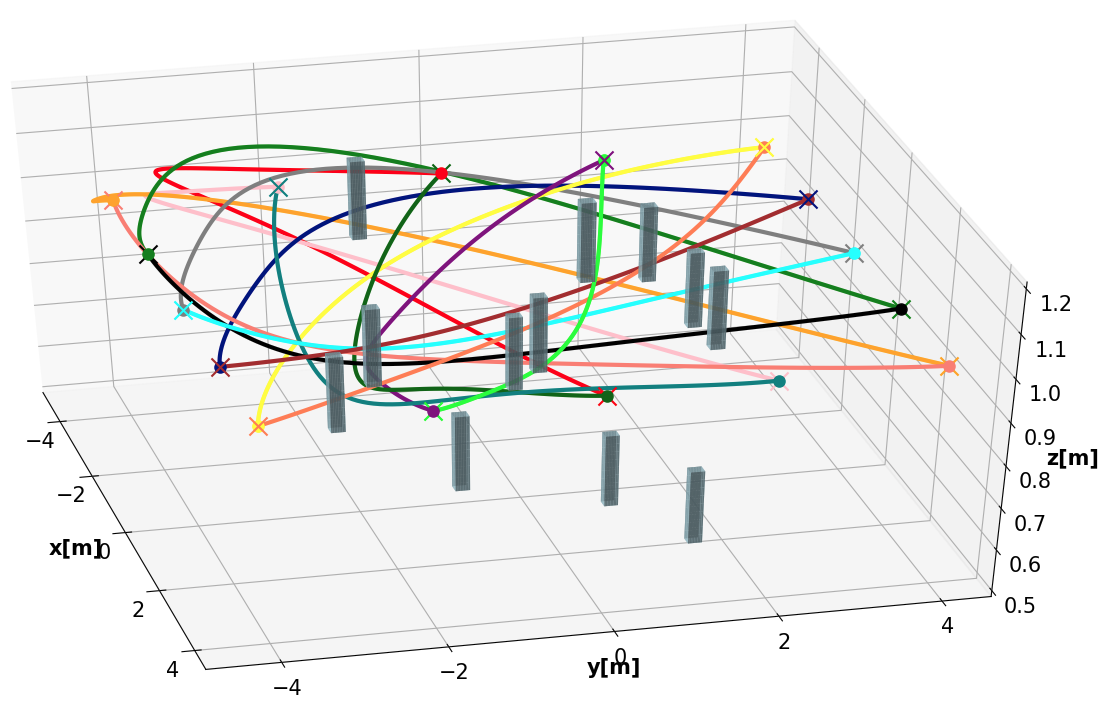}
    \label{random_obstacle_benchmark}
   }
     \subfigure[]{
    \includegraphics[width= 18.0cm, height=5.0cm] {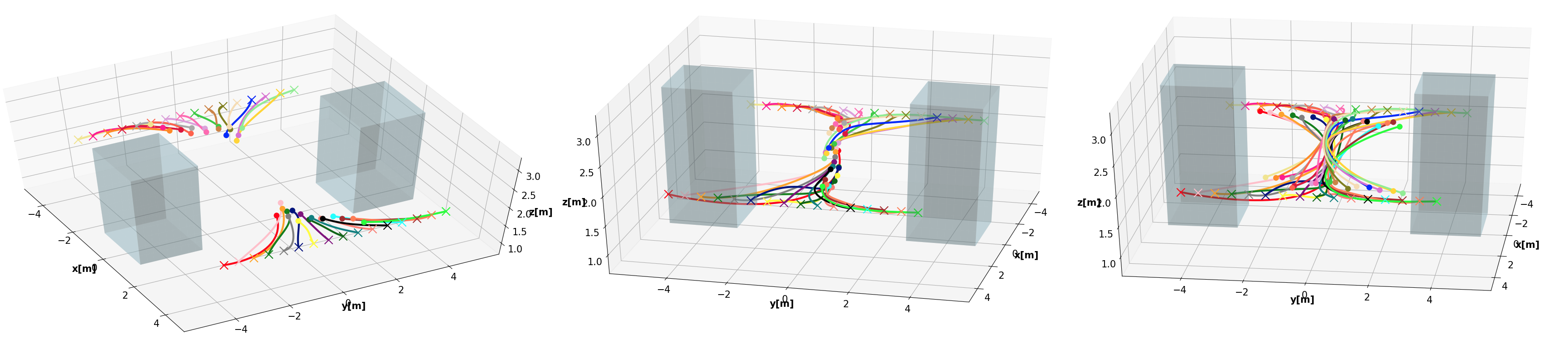}
    \label{16_rob_snap}
   }
\caption{Fig. (a)-(c) shows different benchmarks employed in our experiments along with some typical trajectories obtained with our optimizer. The start and goal positions are marked with a "X" and a "o" respectively. Fig. (d) shows collision avoidance snapshots of 32 agents exchanging positions in a narrow hallway. }
\vspace{-0.4cm}
\end{figure*}

\begin{figure*}[!h]
  \centering  
  \subfigure[]{
    \includegraphics[width= 18.0cm, height=5.0cm] {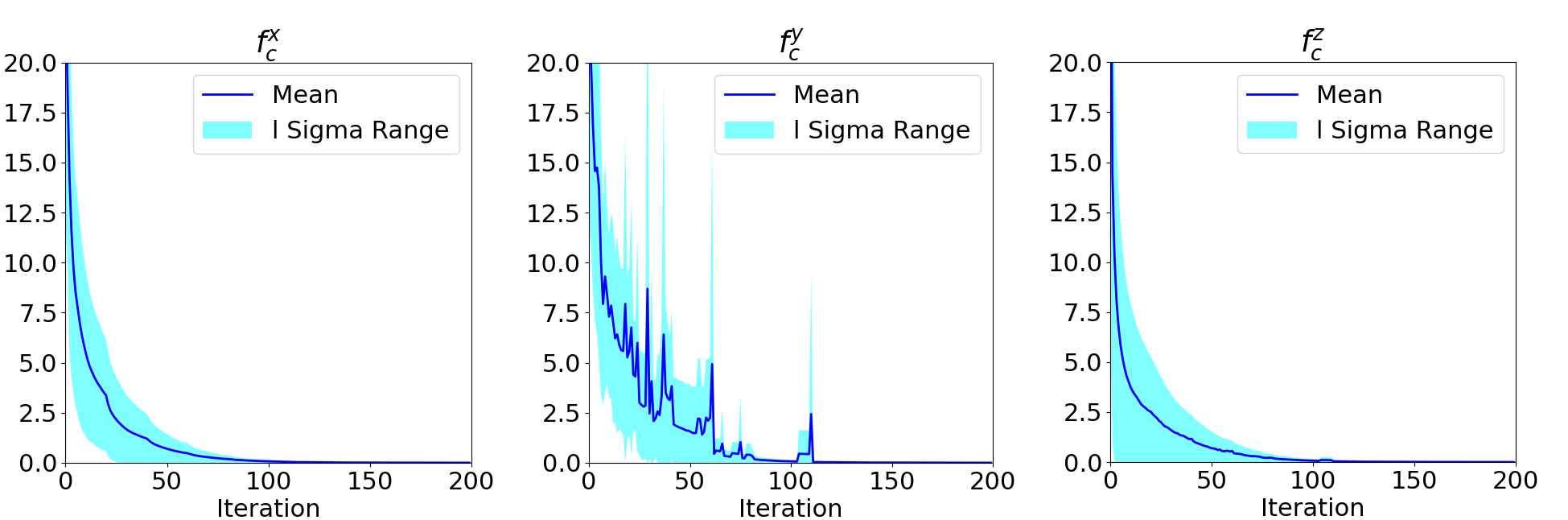}
    \label{residual}
   }\hspace{-0.1cm}
\caption{ The general trend in residual of $\textbf{f}_c$ observed across 20 problem instances. A reliable convergence to zero validates the efficacy of Algorithm \ref{algo_1}. In majority of the benchmarks, a maximum of 150 iterations proved sufficient to get a residual in the order of $10^{-2}$ }
\vspace{-0.5cm}
\end{figure*}

\section{Simulation Results}
\subsection{Implementation Details}
\noindent We implemented Algorithm \ref{algo_1} in Python using CUPY \cite{cupy} and JAX \cite{jax} to accelerate linear algebra on GPUs. Specifically, we use CUPY for trajectory optimization up to 32 agents, while JAX was used for a higher number of agents. Computing hardware consisted of a i7-8750 32 GB RAM desktop computer with RTX 2080 (8GB) and Nividia Jetson TX2.

We pre-computed the inverse of $\widetilde{\textbf{Q}}_x$ in (\ref{kkt_step_x}) for 10 different increasing values of $\rho$ instead of just one. The higher values were used in the latter iterations of Algorithm \ref{algo_1}. This is inspired by \cite{admm_adaptive} that advocates using adaptive $\rho$ to speed up the convergence of optimizer such as Algorithm \ref{algo_1}. It is worth reiterating that for a given number of agents, these inverses are computed only once and can be subsequently used to optimize trajectories from arbitrary start and goal positions. For the ease of implementation, we considered agents as spheres rather than spheroids. Along similar lines, we constructed the static obstacles' circumscribing sphere to incorporate them within our optimizer. For comparison with \cite{rafealla_ccp_quad5} and \cite{rbp_quad}, we used the open-source implementation and data-set provided by the latter. For a fair comparison, we did not include any hard bounds on position, velocities, and accelerations on the implementation of \cite{rafealla_ccp_quad5}. This led to some reduction in the number of inequality constraints. Similarly, we also changed the so called "downwash" parameter in \cite{rbp_quad} to 1 to conform with the implementation of our optimizer. Since trajectories obtained with our optimizer and \cite{rafealla_ccp_quad5}, \cite{rbp_quad} are at different time scales, we used the second-order finite-difference of the position as a proxy for comparing the accelerations across the three methods.

\subsection{Benchmarks and Convergence}
\noindent Fig. \ref{square_benchmark}-\ref{random_obstacle_benchmark} show the typical benchmarks employed in our experiments. In Fig.\ref{square_benchmark}, the agents are placed in a square and are required to navigate to their antipodal positions. In Fig. \ref{random_benchmark}, the start and goal positions are sampled randomly. Fig. \ref{random_obstacle_benchmark} repeats the previous benchmark by adding random static obstacles. Each of the benchmarks was evaluated with a different number of agents with a diverse range of radii. Fig. \ref{16_rob_snap} shows the snapshots of 32 agents exchanging positions in a narrow hallway. Interestingly, our optimizer naturally leads to a line formation pattern among the agents in this benchmark.

A key metric for validating Algorithm \ref{algo_1} is the trend in the residuals of $\textbf{f}_{c}$ (recall (\ref{coll_proposed})) over iterations. It should converge to zero in a diverse set of benchmarks to ensure that the optimizer reliably computes at a collision-free trajectory. Fig.\ref{residual} provides this empirical validation. The plots show the mean and one standard deviation of the residual trajectory obtained across 20 different problem instances. On average, 150 iterations were sufficient to obtain residuals in the order of $10^{-2}$.



\begin{figure}[!h]
  \centering  
    \includegraphics[width= 7.0cm, height=7.0cm] {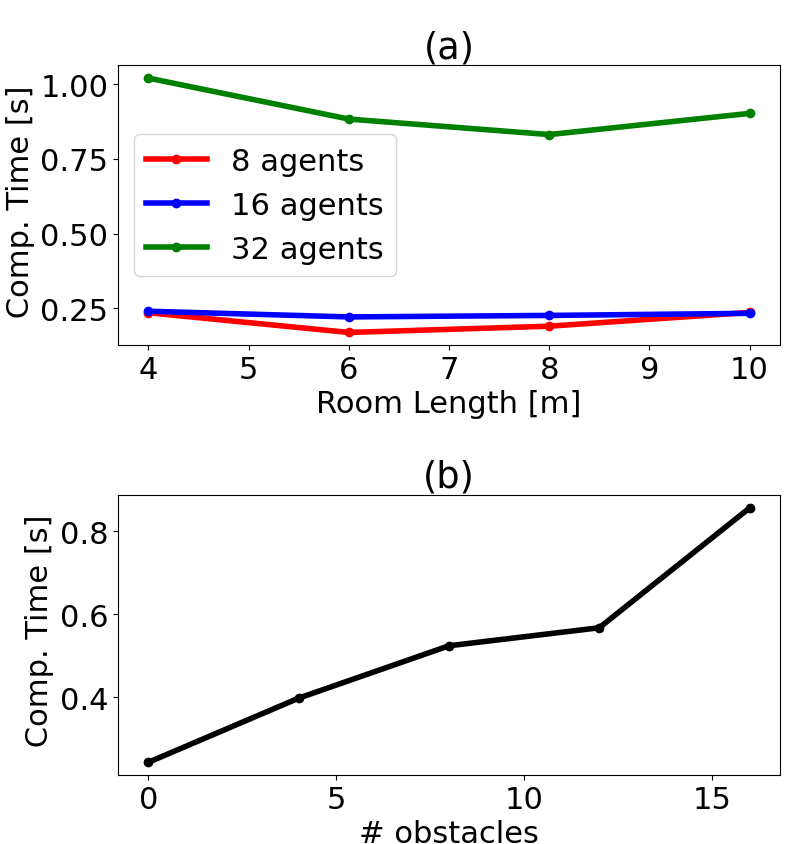}
\caption{ Fig. (a) shows computation time for a varying number of agents
for benchmarks where we sample start and goal positions from a square with varying lengths. Fig. (b) shows the linear scaling of computation time with obstacles for a given number of agents.  }
    \label{comp_time_room}
\end{figure}

\begin{figure}[!h]
    \includegraphics[width= 9.0cm, height=4.0cm] {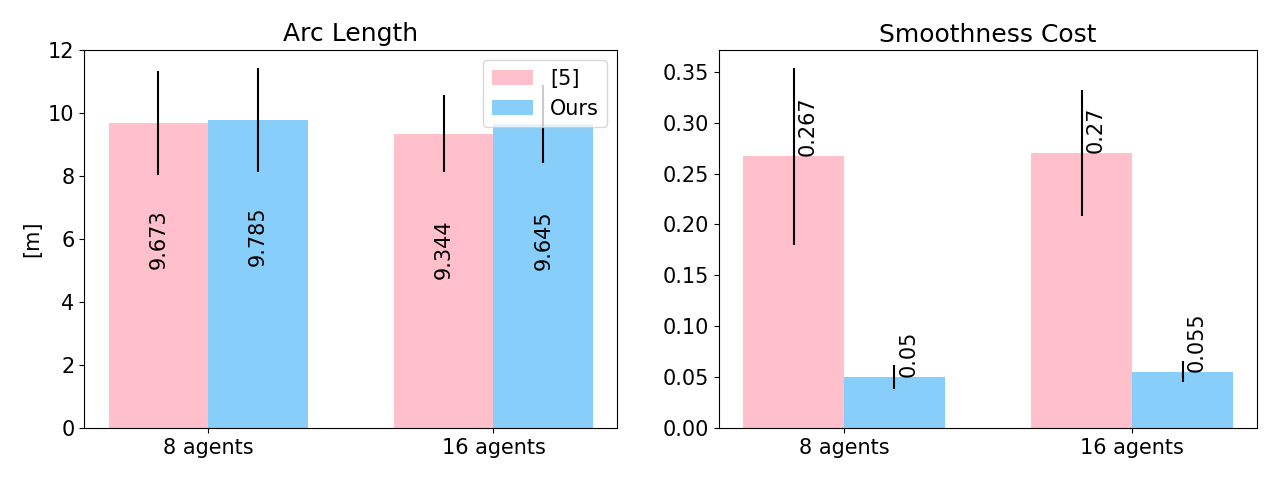}
\caption{Comparison of our optimizer with \cite{rafealla_ccp_quad5} in terms of arc-length and smoothness of the obtained trajectories. The arc-lengths are similar across both the approaches but our optimizer achieves smoother trajectories. Note that the smoothness cost is computed as the norm of the second-order finite-difference of the position of the respective agents at different time instants. }
\label{scp_compare}
\vspace{-0.5cm}
\end{figure}



\subsection{Computation Time}
\noindent Fig. \ref{comp_time_room} (a) presents the mean computation time of our optimizer for a varying number of agents (with radii 0.4 cm) as a function of how closely packed the initial and final positions are. To be more precise, we sampled random initial and final positions in a square room of varying lengths to create problem instances of varying complexity levels. As can be seen, even in the most challenging instance, our optimizer could compute trajectories for 32 agents in around 1s.

Fig. \ref{comp_time_room}(b) shows the computation time for 16 agents with different number of static obstacles. Our optimizer shows an almost linear scaling in computation time. This is because incorporation of static obstacles only affects the computation cost of obtaining $\textbf{A}_{f_c}^T{^k}\textbf{b}_{f_c}^x$ in (\ref{kkt_step_x}). Furthermore, the dimensions of both the matrix and the vector increase linearly with the number of obstacles.

\subsection{Comparison with \cite{rafealla_ccp_quad5} }

\noindent Fig. \ref{scp_compare} presents a comparison of the trajectory quality obtained with our optimizer and \cite{rafealla_ccp_quad5}. Although both the optimizer converge to different trajectories, the arc-length statistics observed across all the agents are very similar. Furthermore, our optimizer outperforms \cite{rafealla_ccp_quad5} in terms of trajectory smoothness cost.
Table \ref{scp_comptime} contrasts the computation time between the two approaches. Our optimizer shows a speed-up of almost 3 orders of magnitude on 8 agent benchmark, and this number shoots up to 60 for the 16 agent benchmark. The computation time trend is not surprising as even state-of-the-art primal-dual interior-point solvers for QP scale cubically with the total number of inequality and equality constraints. Furthermore, the number of inequality constraints stemming from collision avoidance will itself scale as $n \choose 2$. As mentioned earlier, our optimizer by-passes this intractability by pre-computing the expensive matrix inverses and parallelizing matrix-vector product on GPUs. It is essential to point out that GPU accelerations of SCP of  \cite{rafealla_ccp_quad5} is a challenging open problem on its own. We conjecture this to be the motivation behind the same authors adopting the gradient descent approach for leveraging GPUs \cite{quad_gd_gpu}.

\begin{figure*}[!h]
  \centering  
    \includegraphics[width= 18.0cm, height=4.5cm] {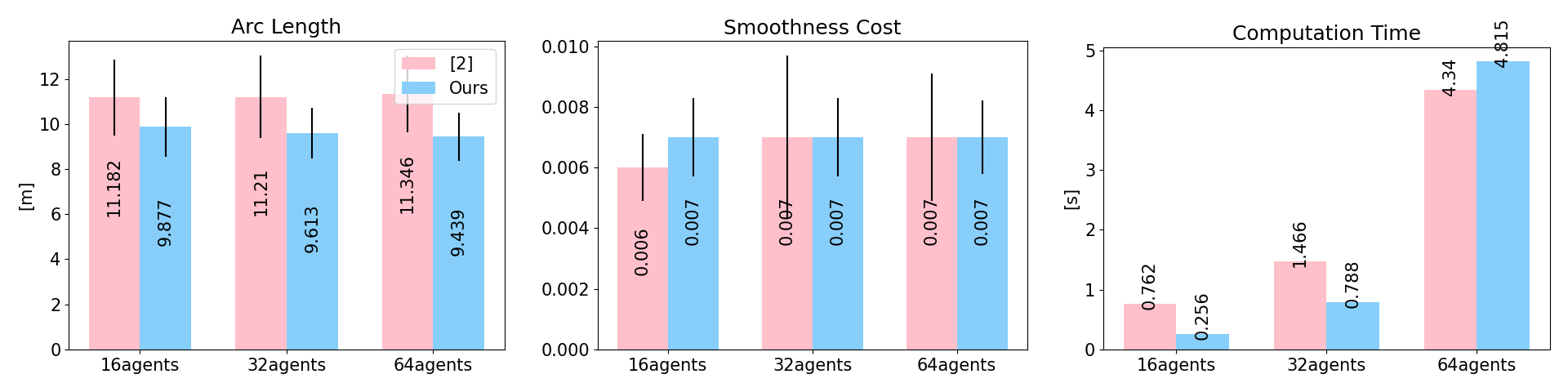}

\caption{Comparisons with current state of the art \cite{rbp_quad}. Our optimizer outperforms \cite{rbp_quad} in terms of trajectory arc-lengths. Even more importantly, it also outperforms in terms of computation time on several benchmarks. }
\label{rbp_compare}
\vspace{-0.5cm}
\end{figure*}

\subsection{Comparison with \cite{rbp_quad}}
\noindent Fig. \ref{rbp_compare} presents the most important result of this paper, wherein we compare our optimizer with the current state of the art \cite{rbp_quad}. The cited work adopts a sequential approach but with a batch of agents. It also leverages the parallel QP solving ability of CPLEX \cite{cplex} on multi-core CPUs. Our optimizer produces trajectories of similar smoothness as \cite{rbp_quad} but with substantially lower arc-lengths. This trend can be attributed to the reduced feasible space accessible to a sequential approach. More surprisingly, our optimizer also outperforms \cite{rbp_quad} in terms of computation time on 16 and 32 agent benchmarks. On the 64 agent benchmark, our optimizer is only marginally slower than \cite{rbp_quad}. We reiterate that it is essential to observe these timings with the context that our optimizer performs a much more rigorous search than \cite{rbp_quad}. The trends in computation time can be understood in the following manner. For a lower number of agents, the computation time of \cite{rbp_quad} is dominated by the niche trajectory initialization it leverages from sampling-based planners. Furthermore, for a lower number of agents, the overhead of CPU parallelization is also significant. But these overheads are easily offset by the computation speed-up achieved for a higher number of agents.

\subsection{Performance on Jetson TX2}

\noindent Table \ref{jetson} shows the computation time for a different number of agents for the square benchmark on Nvidia Jetson TX2. The start and goal positions are sampled from a square of length $8m$. The timings indicate that our optimizer allows for fast on-board decision making for up to 16 agents. Moreover, even for 32 agents, the computation time is small enough to be useful for practical applications. 

\section{Conclusions and Future Work}
In this paper, we fundamentally improved the scalability of joint multi-agent trajectory optimization. Simultaneously, our algorithm is simple to implement and requires the computation of only a few matrix-vector products and element-wise operation over vectors. We achieved this by leveraging hidden geometrical and convex structures in the problem. We outperformed state of the art in joint and sequential approaches in terms of trajectory quality and computation time. One limitation of our optimizer is that it has a constant computational overhead stemming from loading large scale matrices at run-time. However, this limitation has limited practical impact as it needs to be done only once for a given number of agents.  

We are extending our optimizer to work with complicated geometries like an elongated rectangle through multi-circle approximation. Along similar lines, we also aim to extend the results to non-linear agents like autonomous cars by building on bi-convex approximations proposed in our prior work \cite{aks_icra20}.


\begin{table}[h]
\centering
\caption{Computation time comparison with \cite{rafealla_ccp_quad5}}
\label{scp_comptime}
\scriptsize
\begin{tabular}{|l|l|l|}
\hline
 Method & 8 agents  & 16 agents  \\ \hline
Ours & 0.2421  & 0.2682  \\ \hline
 SCP \cite{rafealla_ccp_quad5} & 6.79   & 160.713  \\ \hline
\end{tabular}
\normalsize
\vspace{-0.5cm}
\end{table}

\begin{table}[h]
\centering
\caption{Computation time on Nvidia-Jetson TX2 for Square Benchmark}
\label{jetson}
\scriptsize
\begin{tabular}{|l|l|l|}
\hline
Square Benchmark & Comp. Time [s] \\ \hline
8 agents, radius $= 0.1$ & 1.01  \\ \hline
8 agents, radius $= 0.6$ & 1.320  \\ \hline
8 agents, radius $= 1.2$ & 1.270  \\ \hline
16 agents, radius $= 0.3$ & 2.10  \\ \hline
16 agents, radius $= 0.6$ & 2.34  \\ \hline
32 agents, radius $= 0.25$ & 7.70  \\ \hline
\end{tabular}
\normalsize
\vspace{-0.5cm}
\end{table}




\bibliographystyle{IEEEtran}  
\bibliography{icra_ral_ver1} 

\end{document}